\documentclass[conference]{IEEEtran}
\IEEEoverridecommandlockouts
\usepackage{amsmath,amssymb,amsfonts}
\usepackage{algorithmic}
\usepackage{graphicx}
\usepackage{textcomp}
\usepackage{xcolor}
\usepackage{times}
\usepackage{graphicx}
\usepackage{latexsym}
\usepackage{times}  % DO NOT CHANGE THIS
\usepackage{helvet} % DO NOT CHANGE THIS
\usepackage{courier}  % DO NOT CHANGE THIS
\usepackage[hyphens]{url}  % DO NOT CHANGE THIS
\usepackage{graphicx} % DO NOT CHANGE THIS
\usepackage{graphicx}  % DO NOT CHANGE THIS
\usepackage{tgtermes}
\usepackage{latexsym}
\usepackage{graphicx}
\usepackage{url}
\usepackage{amsmath}
\usepackage{array}
\usepackage{fixltx2e}
\newcolumntype{P}[1]{>{\centering\arraybackslash}p{#1}}
\newcolumntype{M}[1]{>{\centering\arraybackslash}m{#1}}
\usepackage{color, colortbl}
\definecolor{Gray}{gray}{0.9}
\usepackage{soul}
\usepackage{url}
\usepackage[utf8]{inputenc}
\usepackage{amsmath}
\usepackage{microtype}
\usepackage{verbatim}
\usepackage{xcolor}
\usepackage{cleveref}

\usepackage{array}
\usepackage{color, colortbl}
\definecolor{Gray}{gray}{0.9}
\usepackage[utf8]{inputenc}
\usepackage{amsmath}
\newcommand{\cossim}[2]{\mathrm{cossim} \bigg( #1,\, #2 \bigg)}
\newcommand{\vv}[1]{\tilde{v}(#1)}

\usepackage{tikz}
\usetikzlibrary{positioning,shapes.geometric}
\tikzset{
	every node/.append style={
		align=center,
		minimum height=7mm,
	},
	backgroundcolor/.style ={fill=white},
	labe/.append style={
		auto,
		blue,
		align = center,
		font={\itshape}	
	},
}

\usepackage{booktabs, pgfplotstable}
\pgfplotsset{compat=1.16}

\DeclareMathVersion{tabularbold} %see https://tex.stackexchange.com/questions/437144/make-numbers-in-pgfplotstable-bold-without-changing-width
\SetSymbolFont{operators}{tabularbold}{OT1}{ptm}{b}{n} %ptm=times font

\pgfplotstableset{%
	header=has colnames,
	column type={c},
	every head row/.style={before row=\toprule,	after row=\midrule},
	boldcell/.style = {
		postproc cell content/.append style={
			@cell content/.add={\mathversion{tabularbold}}{}
		}
	},
	greycell/.style = {
		postproc cell content/.append style={
			@cell content/.add={\color{gray}}{}
		}		
	},
	integercell/.style = {/pgf/number format/precision=0, column type=r},
	floatcell/.style={/pgf/number format/.cd, fixed, precision=2, fixed zerofill=true},
	floatcellSm/.style={/pgf/number format/.cd, fixed, precision=1, fixed zerofill=true},
	columns/Source/.style={string type, column type=l},
	cortable/.style ={
		columns/Method/.style={string type, column type=r},
		columns/Pearson/.style={floatcell},
		columns/Spearman/.style={floatcell},
		columns/Kendall/.style={floatcellSm},
	},
}

\def\BibTeX{{\rm B\kern-.05em{\sc i\kern-.025em b}\kern-.08em
    T\kern-.1667em\lower.7ex\hbox{E}\kern-.125emX}}
\begin{document}
	
\newcommand{\citet}[1]{\citeauthor{#1} \shortcite{#1}}
\newcommand{\citep}{\cite}
\newcommand{\citealp}[1]{\citeauthor{#1} (\citeyear{#1})}	
%%
%% The "title" command has an optional parameter,
%% allowing the author to define a "short title" to be used in page headers.
\newcommand{\method}{WEmbSim}
\newcommand{\rogue}{ROUGE\textsubscript{L}}
\newcommand{\BLEU}{BLEU\textsubscript{4}}
\newcommand{\methodw}{WEmbSim\textsubscript{W}}
\newcommand{\methodg}{WEmbSim\textsubscript{G}}
\newcommand{\methodf}{WEmbSim\textsubscript{F}}
\newcommand{\methode}{WEmbSim\textsubscript{E}}

\newcommand{\pb}{\textit{R}\textsubscript{BERT-Base}}
\newcommand{\pbm}{\textit{R}\textsubscript{BERT-Base-MRPC}}
\newcommand{\pbl}{\textit{R}\textsubscript{BERT-Large}}
\newcommand{\prb}{\textit{R}\textsubscript{RoBERTa-Base}}
\newcommand{\prl}{\textit{R}\textsubscript{RoBERTa-Large}}
\newcommand{\prlm}{\textit{R}\textsubscript{RoBERTa-Large-MNLI}}
\newcommand{\pxb}{\textit{R}\textsubscript{XLNet-Base}}
\newcommand{\pxl}{\textit{R}\textsubscript{XLNet-Large}}
\newcommand{\pxe}{\textit{R}\textsubscript{XLM-En}}

\title{\method: A Simple yet Effective Metric for Image Captioning\\
}

\makeatletter
\newcommand{\linebreakand}{%
\end{@IEEEauthorhalign}
\hfill\mbox{}\par
\mbox{}\hfill\begin{@IEEEauthorhalign}
}
\makeatother

\author{\IEEEauthorblockN{Naeha Sharif}
	\IEEEauthorblockA{
		\textit{The University of Western Australia}\\
		Perth, Australia \\
		naeha.sharif@research.uwa.edu.au}
	\and
	\IEEEauthorblockN{Lyndon White}
	\IEEEauthorblockA{\textit{Invenia Labs}\\
		 Cambridge, UK \\
		lyndon.white@invenialabs.co.uk}
	\and
	\IEEEauthorblockN{Mohammed Bennamoun}
	\IEEEauthorblockA{\textit{The University of Western Australia}\\
		Perth, Australia \\
		mohammed.bennamoun@uwa.edu.au}
  \linebreakand % <------------- \and with a line-break
	\IEEEauthorblockN{Wei Liu}
	\IEEEauthorblockA{\textit{The University of Western Australia}\\
		Perth, Australia \\
		wei.liu@uwa.edu.au}
	\and
	\IEEEauthorblockN{Syed Afaq Ali Shah}
	\IEEEauthorblockA{
		\textit{Murdoch University}\\
		Perth, Australia \\
		Afaq.Shah@murdoch.edu.au}
	
}

\maketitle

\begin{abstract}
	The area of automatic image caption evaluation is still undergoing intensive research to address the needs of generating captions which can meet adequacy and fluency requirements. Based on our past attempts at developing highly sophisticated learning-based metrics, we have discovered that a simple cosine similarity measure using the Mean of Word Embeddings (MOWE) of captions can actually achieve a surprisingly high performance on unsupervised caption evaluation. This inspires our proposed work on an effective metric WEmbSim, which beats complex measures such as SPICE, CIDEr and WMD at system-level correlation with human judgments. Moreover, it also achieves the best accuracy at matching human consensus scores for caption pairs, against commonly used unsupervised methods. Therefore, we believe that WEmbSim sets a new baseline for any complex metric to be justified. 
\end{abstract}


\section{INTRODUCTION}
Automatic caption evaluation is a challenging task which seeks to score machine generated captions on their quality. Due to a surge of interest in image captioning over the past few years \cite{hossain2019comprehensive}, the development of effective evaluation metrics has become increasingly pressing. Human assessments are considered the gold standard for the evaluation task. However, obtaining human scores for captions is time-consuming and resource intensive. 

Automatic evaluation metrics in contrast are more efficient and cost-effective. However, to serve as a replacement for humans, automatic measures are desired to generate human-like assessments such that their evaluation scores are consistent with human judgments. A number of sophisticated evaluation metrics have been proposed for captioning, amongst which some are borrowed from relevant domains, such as Machine Translation and others are developed specifically for captioning \cite{2017revaluatingcaptioning}, \cite{Sharif2018LearningCompositeMetrics}. While captioning specific metrics such as SPICE \cite{spice2016} and CIDEr \cite{vedantam2015cider} have shown improved performance over the ones adopted from other domains, there is still no consensus on a single metric that gives near human performance.

\begin{figure}[t]
	\centering
	\includegraphics[width=0.45\textwidth]{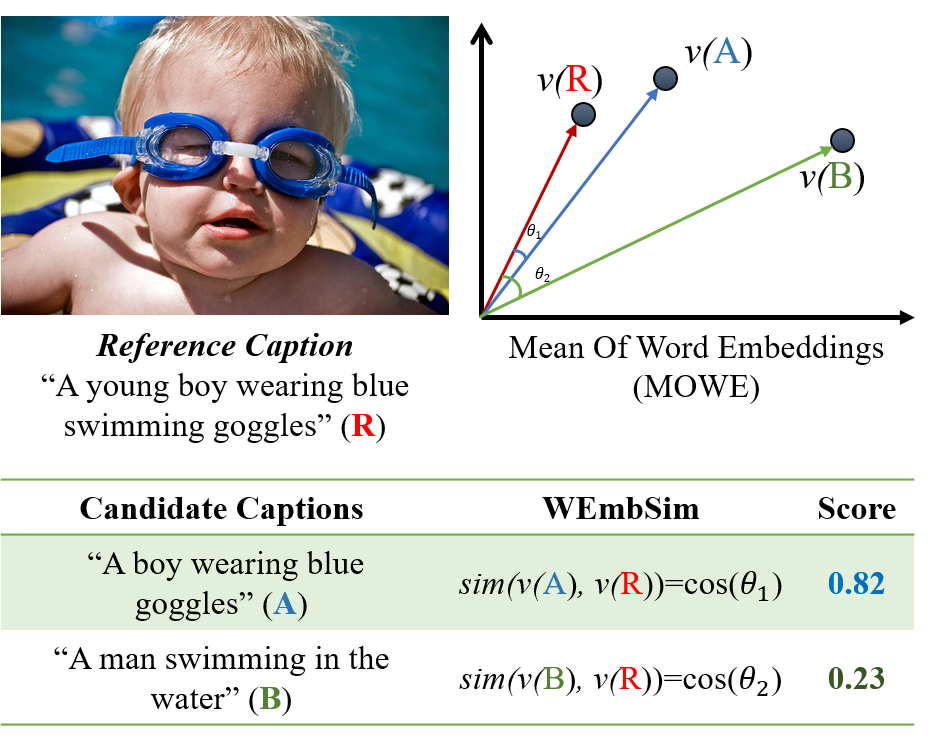}
	
	\caption{An overview of the proposed evaluation metric. The candidate and reference captions corresponding to an image are embedded into a Mean of Word Embeddings (MOWE) space. The evaluation score of a candidate caption is the cosine similarity between its MOWE and that of the reference caption.}
	
	\label{fig:pull up2}
	%\end{center}
\end{figure}

A general inclination of the research community towards sophisticated and intricately crafted measures tends to overlook the efficacy and usefulness of simple methods. To avoid this trap of `complex is better', our research is inspired by the principle of Parsimony (Occam's razor) \cite{sober1981principle}. This principle is interpreted as `a simple solution is better than a complex one'. Thus, if two methods give near equal performance, then one should prefer the simpler one. Our motivation in this work is to propose a simple, intuitive, yet effective metric, that can serve as a strong baseline for caption evaluation, and also be used as a good feature for supervised measures.

In this paper, we present \method{} to evaluate machine generated captions. The key idea is to embed the captions into a semantic embedding space and then to assess the  similarity between caption embeddings. There are various ways to embed sentences in a vector space. Sent2vec \cite{pagliardini2017unsupervised} is one such example, which requires a pretrained model to generate sentence embeddings. In this paper, we adopt the simplest sentence embedding model to represent captions i.e., the representation of captions is obtained through the mean of word embeddings (MOWE). By word embeddings, we mean a family of learned distributed vector representations based on sound language models, such as Word2vec \cite{word2vec} and ELMo \cite{peters2018deep}.

The strength of Word Embeddings in terms capturing semantics has been extensively verified in literature \cite{gladkova2016analogy}, \cite{tshitoyan2019unsupervised}. Moreover, Sums of Word Embeddings (SOWE) have been successfully used as representations in a number of areas \cite{d2019automatic}, \cite{KaagebExtractiveSummaristation}. We show that using the cosine distance between the MOWE of the candidate and reference caption, gives a score that is correlated to human evaluations on the captions quality. To the best of our knowledge, this is the first work which evaluates the performance of such metric for image captioning, using various pre-trained word embeddings across a number of benchmark datasets.

%%%%%%%%%%%%%%%%%%%%%%%%%%%%%%%%%%%%%%%%%%%%%%%%%%%%%%%%%%%%%%%%%%%%%%%%%%%%%%%%%%%%%

For our experiments we make use of pre-trained word embeddings from Word2vec \cite{word2vec}, GloVE \cite{pennington2014glove}, FastText \cite{fasttext} and ELMo \cite{peters2018deep}. We do not use the recently proposed language representation model BERT \cite{devlin2018bert} as it is primarily a fine-tuning based model for applying pre-trained language representations to downstream tasks. Our focus is towards using feature-based strategies for language representations such as ELMo and Word2vec, to avoid the overhead of including the Transformer-based architecture in our caption evaluation task.

Our contributions in this work are as follows: 
\begin{itemize}
	\item We propose a simple yet effective unsupervised metric \method{} for automatic caption evaluation.
	\item To demonstrate the strong performance of \method{} compared to rival methods, we evaluate its system-level correlation and accuracy in terms of matching human consensus scores for caption pairs.
	\item We analyze the robustness of our proposed metric to various sentence perturbations.
	\item We also examine its usefulness and complementarity to the existing measures.  
\end{itemize}

\begin{figure*}[t]
	\centering
	\begin{tikzpicture}[bigstep/.style={draw,minimum height=6cm}]
	\sffamily
	\node(embs)[draw, cylinder, shape border rotate=90, aspect=0.1]{Pre-trained\\Word\\Embeddings};
	
	\node(lookup)[bigstep, below=of embs]{Convert\\Captions\\to\\MOWE \\Representation\\\vspace{0.2cm}\\
		$\vv X=$\\\vspace{0.0cm}\\
		$\dfrac{1}{|X|} \displaystyle \sum_{\forall w_{i}\in X}V_{:,w_{i}}    $
	};
	
	\draw[->, thick] (embs) to (lookup);

	\node[bigstep, right= 1 of lookup](score){Calculate\\Similarity\\\vspace{0.4cm}\\
		$s_i=$ \\\vspace{0.0cm}\\
		$\cossim{\vv {R_i}}{\vv C}$
	};
	
	\node[bigstep, right= 1.5 of score](combine){Combine \\ Scores \\\vspace{0.4cm}\\
		e.g.\\
		$\displaystyle \max_{\forall i} s_i$\\
	};
	
	\node[right= 0.5 of combine](out){$\mathrm{Score}(C \mid \mathcal{R})$};
	\draw[->] (combine) to (out);

	\node[left = 1 of lookup.north west, yshift=-1cm]  (cand) {Candidate Caption};
	\draw[->] (cand.east) to node[labe] {$C$} (cand.east -| lookup.west);
	\draw[->] (cand.east -| lookup.east) to node[labe] {$\vv{C}$} (cand.east -| score.west);
	
	\coordinate[below = 1 of cand] (ref0);
	\def\nref{4}
	\foreach \curr [count=\prev from 0] in {1, ..., \nref} {
		\node(ref\curr)[below=0.1 of ref\prev] {Reference Caption \curr};
		\draw[->] (ref\curr.east) to node[labe] {$R_\curr$} (ref\curr.east -| lookup.west);
		\draw[->] (ref\curr.east -| lookup.east) to node[labe] {$\vv{R_\curr}$} (ref\curr.east -| score.west);
		\draw[->] (ref\curr.east -| score.east) to node[labe] {$s_\curr$} (ref\curr.east -| combine.west);
	};
	
	\node [below = 0.1 of ref\nref, xshift=0.2cm] {
		\color{blue} \emph{Reference Captions} \\
		\color{blue}  $\mathcal{R}=\lbrace R_1, R_2, R_3,R_4 \rbrace$
	};
	
	\end{tikzpicture}
	\vspace{0.5em}
	\caption{Block diagram showing how \method{} is used to score one candidate caption ($C$) against four reference captions $R_1, R_2, R_3, R_4$.}
	\vspace{-0.5em}
	\label{fig:block}
\end{figure*}
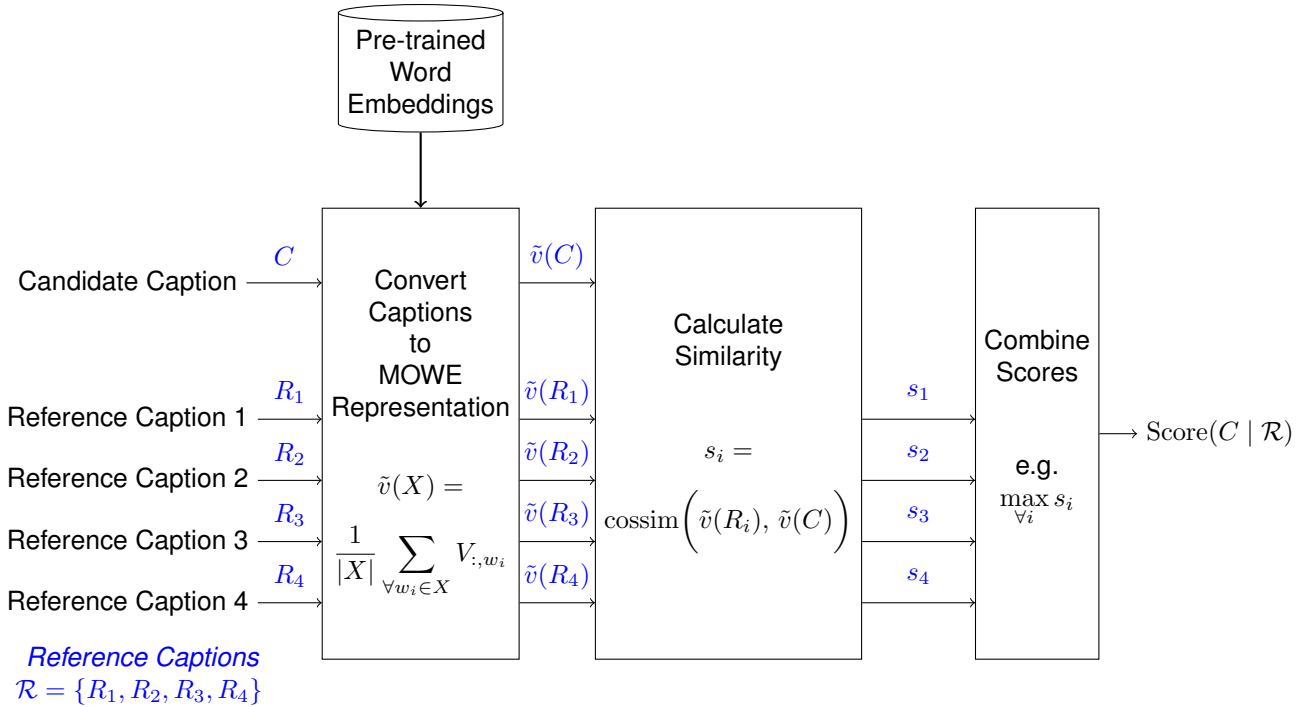

\section{RELATED LITERATURE}

\begin{itemize}
	\item{\textbf{Captioning Vs. Machine Translation:}}
	
	Caption generation has similarities to other natural language generation tasks, in particular machine translation and abstractive summarisation. Image captioning can be thought of as translating images into text. However, not all the contents of an image should be described in the caption: it is expected that the background for example would not be described except if it were particularly relevant to the main action occurring in the image. 
	
	Given the similarities to machine translation and abstract summarisation, it is natural to expect that metrics for those tasks would be useful for the evaluation of captions. However, there are some important differences in the generated text.
	\textit{First}, captions are normally short, single sentences; rather than paragraphs of complex sentences. \textit{Second}, the possible text of captions is restricted by the domain of things that people photograph. While almost any idea can be put in text and translated; there are a wide range of subjects from intangible concepts to non-visually-interesting events, that will never be photographed, and thus never need captions generated for them.
	
	\item{\textbf{Automatic Evaluation Metrics:}}
	
	A number of automatic evaluation metrics have been proposed for captioning, which can be categorized into \textit{supervised} \cite{Sharif2018NNEval}, \cite{lceval}, \cite{cui2018learning} and \textit{unsupervised} \cite{vedantam2015cider}, \cite{spice2016}, \cite{liu2017spider} methods. The work presented here, falls in the latter category. 
	
	\begin{itemize}
		\item{\textbf{Textual Similarity-based Metrics:}}
		Majority of existing caption evaluation metrics capture the linguistic correspondence between the candidate and the reference captions to generate an assessment score. Metrics such as BLEU \cite{Papineni2002}, METEOR \cite{banerjee2005meteor} and CIDEr \cite{vedantam2015cider} measure the n-gram overlap between the candidate and reference captions for evaluation. Such n-gram based metrics rely on word overlaps, therefore, their performance is dependent on the representativeness of the set of reference captions. However, exploiting linguistic information beyond the lexical level, i.e. using pre-trained embeddings helps remedy this issue by extending the reference lexicon. Thus, we believe that metrics which are based on distributed sentence representations are more promising for caption evaluation. 
		
		SPICE \cite{spice2016}, which is a sophisticated semantic metric, estimates the caption quality using a graph-based representation of captions. It relies on semantic parsers and completely ignores the syntactic quality of candidate captions. However, the competitive performance of SPICE shows that measuring the similarities between the semantic representation of captions is useful for caption evaluation. WMD \cite{kusner2015wmd}, which is an embedding-based measure, was originally proposed to measure document similarity. It tends to overly penalize candidate captions which are shorter in length compared to the reference captions \cite{Sharif2018LearningCompositeMetrics}. Moreover, it is also computationally expensive compared to \method{} (Sec.~\ref{subsection: Complexity Analysis}).

		\item{\textbf{Visual Content-based Metrics:}}
		TIGER \cite{jiang2019tiger} and VIFIDEL \cite{madhyastha2019vifidel} are recently proposed metrics which use image content for caption evaluation. TIGEr uses a sophisticated pipeline to project the image (source) and text (target) into a common semantic space for quality assessment. VIFIDEL is an extension of WMD and uses word embedding’s as a bridge for matching content in images and textual descriptions. The performance of TIGEr and VIFIDEL is highly dependent on correct visual detections. VIFIDEL prefers systems that just spit out correct object detections. Thus blurring the line between a list of object detections and captions. 
		
		\item{\textbf{Textual Vs. Visual Content-based Metrics:}}
		Majority of the caption evaluation measures rely on textual similarity to assess the caption quality, presuming that the reference caption sufficiently reflects the visual information. Thus, the quality of references play a crucial role in the case of text-similarity based measures. However, similar is the case with visual evaluation methods such as VIFIDEL which rely on visual detectors to accurately detect the objects. Moreover, comparing visual (source) and textual (target) information is not a straightforward task, and greatly adds to the complexity. 
		
		\item{\textbf{Supervised Metrics:}}
		Supervised metrics such as NNEval \cite{Sharif2018NNEval} and LCEval \cite{lceval} combine existing metrics into a single unified measure, which has shown improvement in performance. However, the drawbacks of learned metrics are their high complexity and subjectivity to the training examples.
	\end{itemize}

	\item{\textbf{Sums of Skip-gram-like Embeddings:}}
	
	The general idea of using sums of skip-gram-like embeddings to represent the meaning of phrases was presented in \cite{mikolovSkip}. FastText, GloVE and Word2vec are all skipgram-like, in that they have a linear relationship between the input embedding and the log of the probability of co-occurring words. Due to this linear-log relationship, a sum of embeddings has an approximately linear-log relationship with the probability of words co-occurring with all the words in the sum (i.e., the intersection of the individual concurrences). This is similar to what would be predicted by a skip-gram-like formulation if the phrase being embedded had been treated as a single token. Thus by the distributional hypothesis (`Words that occur in the same context tend to have similar meaning' \cite{harris1954distributional}), such a sum of words is expected to characterize the sentence meaning.
	
	Authors in \cite{KaagebExtractiveSummaristation} considered a similar approach to measure the sentence similarity as part of their method for extractive summarisation. They found that using sums of embeddings, rather than more complex methods to generate sentence representations gave a superior performance. Authors in \cite{white2015well} and \cite{ritter2015leveraging} found that the idea of using simple mean of embeddings shows strong performances in capturing sentence meaning.
	
	\item{\textbf{Cosine Similarity:}}
	
	The use of the cosine similarity as a measure of correspondence between natural language strings, dates back to the work of \cite{dumais1988LSI} on latent semantic indexing (LSI). This approach was used in information retrieval, to find the most similar documents, based on LSI matrix. LSI vectors are dimensionally reduced word-document occurrence matrices, which often use a weighting that is proportional to with point-wise mutual information (e.g., tdf-if). Skip-gram-like word embeddings are effectively dimensionally reduced word-word occurrence matrices, using a weighting that is proportional to the point-wise mutual information \cite{levy2014EmbMatFact}.
	%\end{enumerate}
\end{itemize}

The efficacy of sums of embeddings as sentence representations has been advocated in a number of research areas \cite{d2019automatic}, \cite{KaagebExtractiveSummaristation}; and the cosine similarity has been very commonly used as a measure \cite{Reisinger2010WSD}, \cite{WhiteRefittingSenses}.  Despite the success of simple mean of embeddings-based measures in other related domains, their effectiveness for image caption evaluation has never been analyzed. To the best of our knowledge, this is the first work which evaluates the effectiveness of such a measure for captioning, across a number of benchmark datasets. 

\begin{table*}[t]
	\centering
	\caption{The details of the pre-trained embeddings used in our investigations.}
	\label{tab:embinfo}
	\begin{filecontents*}{embeddings-info.tsv}
Name 	Source	Dimensions	Corpus	{Corpus Size}	{Vocabulary Size}
%{GloVE 6B 50d}	\cite{pennington2014glove}	50	{Wikipedia + Gigaword} 	6.00E+09	4.00E+05
%{GloVE 42B 300d}	\cite{pennington2014glove}	300	{Common Crawl}	4.20E+10	1.90E+06
{GloVE 840B}	\cite{pennington2014glove}	300	{Common Crawl}	8.40E+11	2.20E+06
{Word2vec}	\cite{word2vec}	300	{Google News (100B)}	1.00E+11	3.00E+06
{FastText}	\cite{fasttext}	300	{Wikipedia}	3.96E+09	2.50E+06
{ELMo} 	\cite{peters2018deep}	1024 {1B Word Benchmark}	1.00E+09	7.93E+05
\end{filecontents*}
\pgfplotstabletypeset[%
	columns/Name/.style={string type, column type=r},
	columns/Corpus/.style={string type},
	columns/{Corpus Size}/.style={column name=\shortstack{Corpus Size}, column type=c, 
		/pgf/number format/.cd, sci, sci zerofill, sci precision=1},
	columns/Dimensions/.style={column name=Dims, integercell},
	columns/{Vocabulary Size}/.style={column name=\shortstack{Vocabulary Size}, integercell , column type=c},
	]{embeddings-info.tsv}
	
\end{table*}

An unpublished concurrent work \cite{bert-score} uses pre-trained BERT embeddings and token-level computations to evaluate the similarity between two captions. In contrast, our metric is based on sentence-level representations, and is comparatively simpler. We focus on using feature-based strategies for language representations such as ELMo and Word2vec to avoid the overhead of including the Transformer based encoder architecture, which is used by BERTScore. Moreover, what we are proposing is a framework, which is not confined to a particular embedding technique.

\section{THE PROPOSED METRIC}
\method{} is a cosine similarity based measure which uses the mean of word embeddings
for caption evaluation. \method{} is defined mathematically as follows. For a set of reference captions $\mathcal{R}$,
and a candidate caption to be evaluated $C=\left[w_1, w_2, \ldots, w_n\right]$, we define the function $\tilde{v}(\cdot)$ which maps a caption to a vector 
via the mean of word embeddings representation using the embedding matrix $V$.
\begin{equation}
\vv C=\frac{1}{n}\sum_{\forall w_{i}\in C}V_{:,w_{i}}    
\end{equation}
Using this, and the cosine similarity function:
\begin{equation}
\cossim{\tilde a}{\tilde b} = \dfrac{|\tilde a \cdot \tilde b|}{|\tilde a||\tilde b|}
\end{equation}

We define the scoring function as:
\begin{equation}
\mathrm{Score}(C \mid \mathcal{R})
= \underset{\forall R_{i} \in \mathcal{R}}{\mathrm{Combine}}\, 
\cossim{\vv C}{\vv {R_{i}}}% 
\end{equation}

The scoring function is defined in terms of the \textit{combining rule}.
The combining rule specifies how to combine the scores for multiple references.
The similarity is defined between two captions, and as in caption evaluation usually multiple references are provided, thus the scores must be combined.
It is standard in other similar methods, such as METEOR \cite{banerjee2005meteor} to use \textit{max} as the combining function \cite{banerjee2005meteor}.
This corresponds to being generous with the score i.e., it is acceptable to be at least similar to one of the reference captions.

The alternative approach is to use \textit{min} which requires the candidate to be similar to all of the reference captions since it reports the worst score. In contrast, using the \textit{mean} strikes a balance between the two. We investigated all three i.e., \textit{max}, \textit{mean} and \textit{min}, however, for brevity, we report the results for using the \textit{mean} combining function for \method{}, as \textit{mean} shows consistently better performance than either \textit{min} and \textit{max}.

%%%%%%%%%%%%%%%%%%%%%%%%%%%%%%%%%%%%%%%%%%%%%%%%%%%%%%%%%%%%%%%%%%%%%%%%%%%%%%%%%%%%%%%

\subsection{Pre-trained Embeddings}
\label{subsection: Pre-trained Embeddings}
We use off-the-shelf pre-trained embeddings, without any fine tuning for this task.
The details of the embeddings we used in our investigations are given in Table~\ref{tab:embinfo}. For ELMo, we use the representations from the lowest layer of BiLM \cite{peters2018deep}, as they show consistently better performance on our test datasets, compared to the other two higher layers or a linear combination of the three layers. As discussed in \cite{peters2018deep}, the lower level LSTM states capture syntactical aspects of a sentence. For our captioning datasets, syntactic information proved to be more useful in differentiating between captions. We evaluate \method{} using different embedding to assess how various embeddings affect its performance.

\subsection{Preprocessing}
We perform preprocessing of the captions to convert them to a MOWE representation.
We remove stop words, using the standard NLTK \cite{NLTK}
stop-word list.
We found in preliminary investigations that removing stop-words gives a small improvement across the board for all embeddings.
We also remove any words that are outside the pre-trained vocabulary for the embeddings.
We conducted some investigations into using char-ngram embeddings for out-of-vocabulary 
words; we however achieved only a very small improvement, as rare and stop words contribute little to the performance. Therefore, for simplicity we do not use these methods in our experiments.

\subsection{Complexity Analysis}
\label{subsection: Complexity Analysis}
Our proposed metric \method{}, superficially sounds similar to the Word Mover's Distance (WMD) proposed by \cite{kusner2015wmd}. Both methods use word embeddings to compute the similarity of natural language. However, WMD is a much more complex method, as it attempts to attribute the distance at the word level. Whereas, our proposed method \method{}, just uses the mean of the word embeddings to embed sentences and measures the similarity between sentence embeddings. WMD has time complexity per pairwise evaluation $O(p^3 \log p)$, in contrast per pairwise evaluation time complexity of \method{} is \textbf{ $O(p^2)$}, where $p$ is the length of the captions. \method{} is much more similar to the lower bound that \cite{kusner2015wmd} places on WMD, which is referred to as the word centroid distance. 
%%%%%%%%%%%%%%%%%%%%%%%%%%%%%%%%%%%%%%%%%%%%%%%%%%%%%%%%%%%%%%%%%%%%%%%%%%%%%%%%%%%%%%%%%%%%%%

\begin{table}
	\begin{center}
		\caption{Pearson's $p$ correlation of commonly used evaluation metrics against 12 competition entries on the COCO validation set \cite{cui2018learning}. Our reported scores with a * differ from the ones reported in \cite{cui2018learning} M1: Percentage of captions that are evaluated as better or equal to human caption.
			M2:	Percentage of captions that pass the Turing Test. Highest correlation values are shown in boldface.
			\label{table:system-level}}
		\begin{tabular}{m{1.4cm}M{1.1cm}M{1.1cm}M{1.1cm}M{1.1cm}}
			\hline
			& \multicolumn{2}{c}{M1} & \multicolumn{2}{c}{M2}  \\
			\hline
			Metric & $p$ & p-value &   $p$ &  p-value \\ \hline
			\BLEU{}	&	-0.02 &	0.95 &	-0.01 &	0.99 \\
			\rogue{} &	0.09 &		0.77 &	0.10 &	0.75\\
			METEOR	&	0.61 &		0.03 &	0.59 &	0.03  \\
			CIDEr &	0.47* &		0.13* &	0.50* &	0.10* \\
			SPICE	&	0.76* &		0.00* &	0.75* &	0.01* \\
			WMD 	&	0.68 &	0.02 &	0.71 &	0.01 \\
			\methode{}&	0.47 &	0.13 &	0.48 &	0.12\\
			\methodg{}	&	0.80 &		0.00 &	\textbf{0.76} &	0.00 \\
			\methodw{}	&	0.71 &		0.01 &	0.64 &	0.02 \\
			\methodf{}	&	\textbf{0.83} &		0.00 &	\textbf{0.76} &	0.01 
			\\ \bottomrule
		\end{tabular}
	\end{center}
\end{table}

\section{EXPERIMENTS AND RESULTS}
\label{section: EXPERIMENTS AND RESULTS}
In this section we present the details of the four different experiments that we performed to validate the effectiveness of our proposed metric. Since our metric falls in the category of \textit{unsupervised textual-comparison metrics}, we compare its performance to other commonly used image captioning metrics, which belong to the same category.

%%%%%%%%%%%%%%%%%%%%%%%%%%%%%%%%%%%%%%%%%%%%%%%%%%%%%%%%%%%%%%%%%%%%%%%%%%%%%%%%%%%%%
%%%%%%%%%%%%%%%%%%%%%%%%%%%%%%%%%%%%%%%%%%%%%%%%%%%%%%%%%%%%%%%%%%%%%%%%%%%%%%%%%%%%%
\subsection{System-level Correlation}
\label{subsection:System-level Correlation}

Strong correlation with human judgments is considered one of the most important qualities of an automatic evaluation metric \cite{zhang2010significanceofMTmetrics}. We investigate the system-level correlation of our proposed metric against the human assessments. Since image captioning metrics are used to evaluate captioning systems, it is therefore important to compare metrics in terms of their system-level performance.  

For this experiment we use the system-level manual evaluations of 12 teams that participated in MSCOCO Captioning Challenge and submitted their results on the validation set. Similar to \cite{cui2018learning}, we assume that the manual assessments on the validation and test set are fairly similar. In Table~\ref{table:system-level} we report Pearson's $p$ against  M1 and M2 human assessments, which were used to rank the captioning models. Where, M1 is the `percentage of captions that are evaluated as better or equal to human captions' and M2 is the `percentage of captions that pass the Turing test'.

In Table~\ref{table:system-level}, we report four variants of our metric; \methode{}, \methodg{}, \methodf{} and \methodw{}, which use ELMo, GloVe, FastText and Word2vec respectively. 
Table~\ref{table:system-level} shows that our best performing metric \methodf{} significantly outperforms all exsisting handcrafted metrics, achieving a correlation of 0.834 and 0.755 with human judgments M1 and M2 respectively.

%%%%%%%%%%%%%%%%%%%%%%%%%%%%%%%%%%%%%%%%%%%%%%%%%%%%%%%%%%%%%%%%%%%%%%%%%%%%%%%%%%%%%%

%%%%%%%%%%%%%%%%%%%%%%%%%%%%%%%%%%%%%%%%%%%%%%%%%%%%%%%%%%%%%%%%%%%%%%%%%%%%%%%%%%%%%%

%%%%%%%%%%%%%%%%%%%%%%%%%%%%%%%%%%%%%%%%%%%%%%%%%%%%%%%%%%%%%%%%%%%%%%%%%%%%%%%%%%%%%%
\subsection{Pairwise Classification Accuracy}

We also evaluate the accuracy of our proposed and several other key metrics at matching human consensus score for candidate captions. For this experiment we use PASCAL-50S and ABSTRACT-50S \cite{vedantam2015cider} datasets, which contain human consensus judgements for pairs of captions. The human scores for the candidate captions were collected through AMT, where the workers were shown three captions, two candidates and one reference, and were asked to pick the candidate which is most similar to the reference caption.  This choice was based solely on sentence similarity, as the workers were never shown the corresponding image. PASCAL-50S contains 50  whereas ABSTRACT-50S contains 48 reference captions per candidate pair.

For our experiments we focus on comparing two human written captions i.e., 1) Human Human Correct (HHC) and \textbf{2}) Human Human Incorrect (HHI). In HHC and HHI categories, both captions are human written, where in HHC both are correct and in HHI one candidate caption is incorrect. PASCAL-50S also contains caption pairs generated by machine models (MM category). However, we notice that majority of the machine generated captions are of inferior quality, as the image captioning systems used for generating them are retrieval-based, rule-based or template-based \cite{vedantam2015cider}. Captions generated by such models are not representative of the state-of-the-art. Therefore, we only used captions from HHC and HHI category, as they are comparatively closer in quality to the captions generated by state-of-the-art models. We compute the scores for the given candidate pairs using five randomly selected references, as done in literature \cite{spice2016}. Ideally, a metric should assign a higher score to the captions preferred by humans.

\begin{table}
	\begin{center}
		\caption{Pairwise classification accuracy on two different tasks reported on ABSTRACT-50S and PASCAL-50S \cite{vedantam2015cider}. The highest accuracy for each task is shown in bold face.}
		\label{tab:results-abstract}
		\begin{tabular}{@{}lccccc@{}}
			\toprule
			\multicolumn{1}{l}{Metric}        & \multicolumn{2}{l}{ABSTRACT-50S}  & \multicolumn{2}{l}{PASCAL-50S} & \multicolumn{1}{l}{Avg.}              \\ \midrule
			{}                      & HHC            & HHI            & HHC           & HHI           & {}                              \\
			\hline
			\BLEU{}                        & 51.5           & 81.0              & 53.7          & 93.2          & 69.9                              \\
			\rogue{}   					& 52.0             & 82.0             & 56.5          & 95.3          & 71.5                              \\
			METEOR                      & 48.5           & 46.5           & \textbf{61.1} & 97.6          & 63.4                              \\
			CIDEr                       & 50.5           & 78.5           & 57.8          & 98.0          & 71.2                              \\
			SPICE                       & 56.0             & 86.5           & 58.0          & 96.7          & 74.3                              \\
			WMD                         & 50.0             & 93.5           & 56.2          & 98.4          & 74.5                              \\
			\methode{} 					& 55.5           & \textbf{94.5}  & 56.5          & 98.3          & 76.2                              \\
			\methodg{} 					& 56.0             & 89.5           & 58.0          & 97.6          & 75.3                              \\
			\methodw{} 					& 57.5           & 90.5           & 59.3          & \textbf{98.9 }         & 76.6                              \\
			\methodf{} 					& \textbf{60.5}  & 90.0             & 60.9          & 98.1          & \textbf{77.4}\\
			\hline
			\# Instances  				&1000			&1000		&1000			&1000 		&4000\\ \bottomrule
		\end{tabular}
	\end{center}
\end{table}

\begin{figure*}[t]
	\centering
	\includegraphics[width=0.75\textwidth]{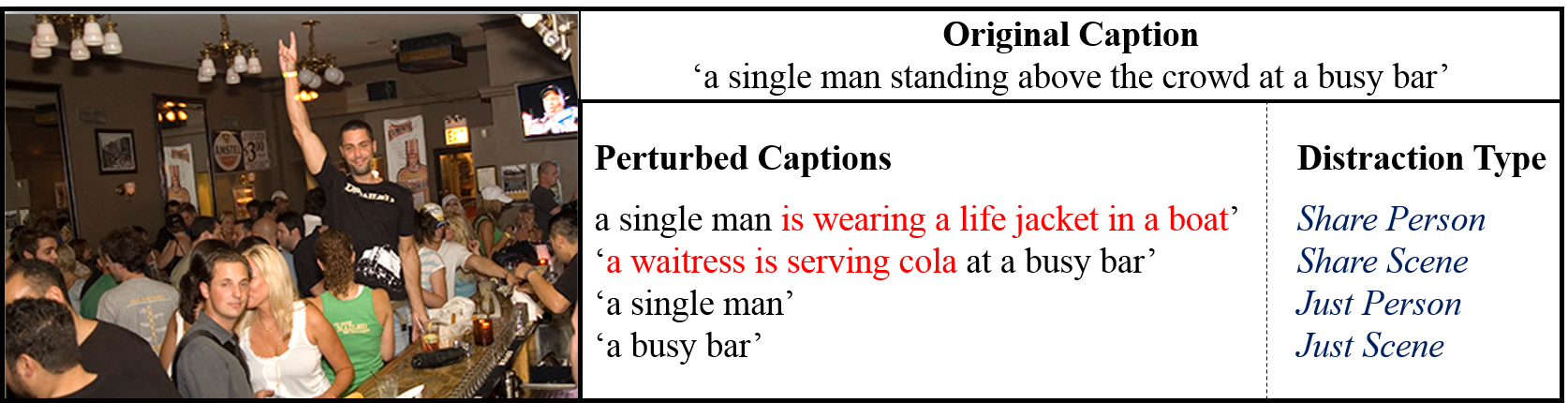}
	\caption{shows an image, corresponding correct caption, perturbed versions of the original caption and the perturbation types.}
	\label{fig:Distraction Example}
	%\end{center}
\end{figure*}

Table~\ref{tab:results-abstract} shows the results of this experiment. In HHC, most of the metrics achieve a low accuracy since differentiating between good quality captions is a difficult task. However, \methodf{} achieves the highest accuracy of 60.5 on ABSTRACT-50S and comes very close (60.9) to the best performing metric METEOR (61.1) on PASCAL-50S. All other variants of our proposed measure, show a higher overall accuracy compared to other handcrafted measures. The results in Table~\ref{tab:results-abstract} highlight that \method{}, is a strong baseline at matching human consensus scores for captions which differ in quality.

\subsection{Distraction Analysis}

We also analyse the robustness of our proposed metric against various sentence perturbations. For this experiment we use the dataset introduced in \cite{hodosh2016focused}. We report the performance on four different tasks namely \textbf{1}) Share Person (SP), \textbf{2}) Share Scene (SS), \textbf{3}) Just Person (JP) and \textbf{4}) Just Scene (JS). An example of each of the four tasks is shown in Figure~\ref{fig:Distraction Example}. For the SP and SS tasks, the distractors contain the same person/scene as the correct caption, however, the remaining part of the sentence is different. The JS and JP distractors are just noun phrases and consist only of the scene/person of the correct caption.

\begin{table}
	\begin{center}
		\caption{The distraction analysis on binary forced-choice tasks. The highest accuracy for each task is shown in bold face.}
		\label{tab:results-distract}
		\begin{tabular}{@{}lccccc@{}}
			\toprule
			Metric                      & \begin{tabular}[c]{@{}c@{}}Share\\ Person\end{tabular} & \begin{tabular}[c]{@{}c@{}}Share\\ Scene\end{tabular} & \begin{tabular}[c]{@{}c@{}}Just\\ Person\end{tabular} & \begin{tabular}[c]{@{}c@{}}Just\\ Scene\end{tabular} & Avg.          \\ \midrule
			\BLEU{}                        & 83.5                                                   & 82.4                                                  & 54.9                                                  & 67.7                                                 & 72.1          \\
			\rogue{}   & 86.8                                                   & 86.8                                                  & 83.4                                                  & 94.1                                                 & 87.8          \\
			METEOR                      & 92.4                                                   & 91.4                                                  & 91.9                                                  & 98.4                                                 & 93.5          \\
			CIDEr                       & 94.1                                                   & \textbf{93.1   }                                               & 73.3                                                  & 81.5                                                 & 85.5          \\
			SPICE                       & 88.5                                                   & 88.8                                                  & 78.1                                                  & 92.0                                                 & 86.9          \\
			WMD                         & \textbf{94.3}                                                   & 92.5                                                  & 86.3                                                  & 96.9                                                 & 92.5          \\
			\methode{} & 92.5                                                   & 91.4                                                  & 89.4                                                  & 96.6                                                 & 92.5          \\
			\methodg{} & 91.8                                                   & 90.2                                                  & 95.1                                                  & 99.6                                                 & 94.2          \\
			\methodw{} & 93.9                                                   & 92.2                                                  & 93.6                                                  & 98.8                                                 & 94.6          \\
			\methodf{} & 92.5                                                   & 91.2                                                  & \textbf{95.7}                                         & \textbf{99.6}                                        & \textbf{94.8} \\
			\hline
			\# Instances                & 4596                                                   & 2621                                                  & 5811                                                  & 2624                                                 & 15,652
			\\ \bottomrule
		\end{tabular}
	\end{center}
\end{table}

A robust metric should assign a higher score to the correct caption as compared to its perturbed version. Table~\ref{tab:results-distract} shows that \methodf{} achieves the highest accuracy in JP and JS category, showing that it can identify when a complete sentence (unperturbed) is better than just a noun phrase (distractor). It can be noted, that other sophisticated metrics are comparatively not that robust to such perturbations. We also notice that the embedding-based measures such as WMD and the variants of our \method{} achieve a higher overall accuracy on the four tasks compared to the other measures. 
\begin{figure*}[t]
	\centering
	\includegraphics[width=0.75\textwidth]{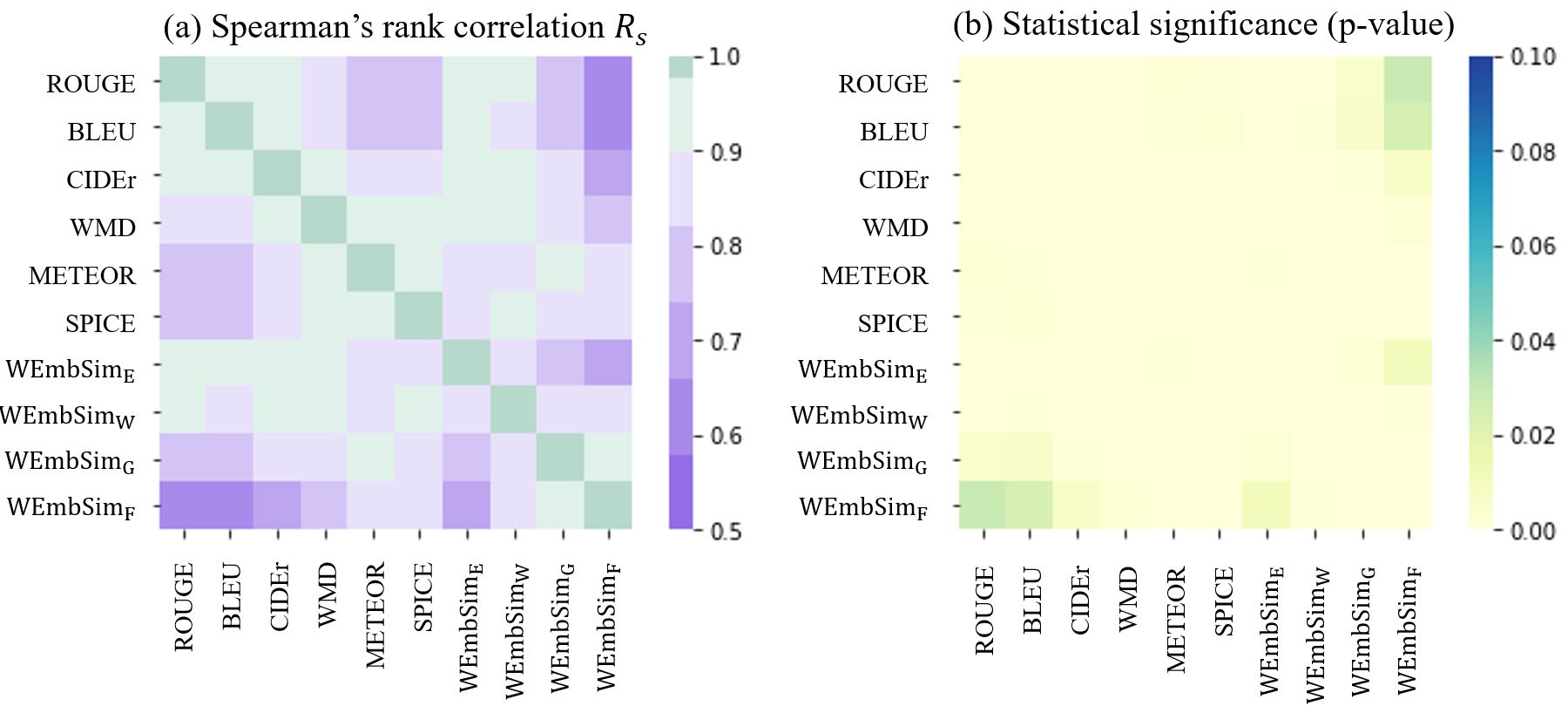}
	\caption{Pairwise Spearman correlation of automatic metrics on MSCOCO captioning challenge validation set.}
	\label{fig:spearman Example}
	%\end{center}
\end{figure*}

\begin{table*}[t]
	%\resizebox{\textwidth}{!}{%
	\centering
	\caption{Pearson correlation between standalone/combined metrics and human assessments for 12 teams on MSCOCO validation set.}
	\label{tab:my-table}
	\begin{tabular}{m{1.1cm}M{2cm}M{2cm}M{2cm}M{2cm}M{2cm}}
		\toprule
		& Metric & \begin{tabular}[c]{@{}l@{}}Metric +\\\methode{}\end{tabular} & \begin{tabular}[c]{@{}r@{}}Metric + \\\methodw{}\end{tabular} & \begin{tabular}[c]{@{}r@{}}Metric + \\\methodg{}\end{tabular} & \begin{tabular}[c]{@{}l@{}}Metric + \\\methodf{}\end{tabular} \\ \midrule
		\BLEU{} & 0.529  & 0.520                                                   & 0.609                                                 & 0.613                                                   & \textbf{0.652}                                             \\
		\rogue{}  & 0.388  & 0.411                                                   & 0.519                                                 & 0.522                                                   & \textbf{0.567}                                             \\
		METEOR & 0.744  & 0.658                                                   & 0.748                                                 & 0.778                                                   & \textbf{0.811}                                             \\
		WMD    & 0.682  & 0.618                                                   & 0.712                                                 & 0.736                                                   & \textbf{0.773}                                             \\
		CIDEr  & 0.465  & 0.466                                                   & 0.500                                                 & 0.500                                                   & \textbf{0.515}                                             \\
		SPICE  & 0.762  & 0.656                                                   & 0.745                                                 & 0.785                                                   & \textbf{0.813}                                            
	\end{tabular}
	
\end{table*}

Our metric performs no direct assessment of the word order and primarily focuses on the word content (semantics). Similar to SPICE, \method{} neglects fluency of captions as compared to other n-gram based measures. An obvious weakness of \method{} is that it can trivially be fooled if it is presented with a caption that has the right content, in a random order. For example, one could create an unordered list of keywords generated using object detection. This would be scored quite highly by \method{}. 
The metric is thus  `gameable'. However, if this were expected, such inputs could be readily detected and assigned a poor score either by using a general purpose language model, or by setting a minimal threshold on the BLEU score.

\subsection{Is \method{} complementary to other commonly used captioning measures?}

%Our proposed method performs no direct assessment of the word order. Therefore, in order to make it robust to gameability, \method{} can be combined with a measure which evaluates fluency. An example would be to set a minimal threshold on the BLEU score.

%%%%%%%%%%%%%%%%%%%%%%%%%%%%%%%%%%%%%%%%%%%%%%%%%%%%%%%%%%%%%%%%%%%%%%%%%%%%%%

% Please add the following required packages to your document preamble:
% \usepackage{booktabs}

To elucidate the usefulness and complementarity of \method{} to the existing unsupervised metrics, we compute the correlation between our proposed measure and other existing metrics. For this experiment, we use the same dataset that we used for the system-level evaluation  (Sect.~\ref{subsection:System-level Correlation} for details). We report Spearman correlation coefficient to maintain consistency with the literature \cite{2017revaluatingcaptioning}. From Figure~\ref{fig:spearman Example} we observe that the correlation between metrics that capture the similar linguistic properties is higher. For example lexical measures such as BLEU and \rogue{} correlate very strongly with each other compared to semantic measures such as SPICE and WMD. Correlation of METEOR is relatively lower to lexical measures compared to the semantic ones, since it captures the semantic information using synonym matching. Amongst all measures \methodf{} shows a relatively lower correlation with the existing measures and even with the two other variants \methode{} and \methodw{}. This shows that \methodf{} is complementary to the existing measures. 

To further strengthen our claim,  we linearly combine the scores of the commonly used measures and the variants of \method{} for 12 teams that participated in MSCOCO Captioning Challenge (Sec. 4.1 for details). We evaluate and report the Pearson correlation of the combined scores with the human judgements in Table~\ref{tab:my-table}. It can be seen from Table~\ref{tab:my-table} that combining \methodf{} to the existing measures significantly improves their performance. This reflects the complementarity and effectiveness of our proposed measure.

%%%%%%%%%%%%%%%%%%%%%%%%%%%%%%%%%%%%%%%%%%%%%%%%%%%%%%%%%%%%%%%%%%%%%%%
\section{DISCUSSION}

The quality of the generated text can be rated on its \textit{fluency} and \textit{adequacy} \cite{Sharif2018LearningCompositeMetrics}.
\textit{Fluency} is the quality of the text to flow-well, and have both natural, and grammatically correct word order.
\textit{Adequacy} is the quality of containing the correct and suitable information. A vast majority of the state-of-the-art captioning models \cite{hossain2019comprehensive} are based on Recurrent Neural Network (RNN) \cite{mikolov2010rnnlm} language models for text generation. RNN models perform exceedingly well at generating text with the correct order \cite{mikolov2010rnnlm}. We thus can assume that the generated captions are usually fluent. Therefore, the meaningful composition of the words becomes a better discriminator in assessing the caption quality. Thus checking the word content takes priority over the word order.

Linear combination of word embeddings (mean,sum) have shown to be surprisingly powerful at the semantic tasks \cite{arora2016simple}.
This is explained in large by their ability to capture how words are used
\cite{acl2018probingsentencevectors},
and a certain degree of lexical similarities, such as synonyms
\cite{mikolov2013linguisticsubstructures,pennington2014glove}.
If one has the correct word content, even a basic tri-gram language model is often able to produce the correct sentence, particularly for short sentences \cite{White2016BOW2sent}. Captions are normally short, single sentences; rather than paragraphs of complex sentences. The simple linear combination of word embeddings has shown its ability in capturing the compositional semantics \cite{wang2017matrix}, which confirms our finding on its usefulness for caption evaluation in this work.

Our experiments reflect the significance of the choice of embeddings in our proposed model. FastText achieved the most promising results compared to other word embeddings. The difference in performance can be attributed to various factors such as the word semantics encoded by the pre-trained word embeddings, training criteria and training data. However, as the word embedding models will continue to evolve, the performance of our proposed metric can be improved even further by using stronger, effective and relevant embeddings. Note however, the purpose of this work is not to propose the best metric, but rather setting up a hard-to-beat simple baseline for more complex metrics to be justified.  
%%%%%%%%%%%%%%%%%%%%%%%%%%%%%%%%%%%%%%%%%%%%%%%%%%%%%%%%%%%%%%%%%%%%%%%%%%%%

%%%%%%%%%%%%%%%%%%%%%%%%%%%%%%%%%%%%%%%%%%%%%%%%%%%%%%%%%%%%%%%%%%%%%%%%%%%%
\section{CONCLUSION}
We have presented a new metric \method{} to automatically evaluate captions.
\methodf{}, using FastText set of embeddings achieves the highest system-level correlation against existing unsupervised methods and, it also achieves state-of-the-art results on the pairwise classification task (HHI) on ABSTRACT-50S. A comprehensive correlation study against popular unsupervised metrics shows that \method{} is complementary to other commonly used measures.

Our work highlights the surprising effectiveness of word embeddings and \method{} can serve as a strong baseline for caption evaluation. This is in the same sense of a null model to any machine learning model. In order to justify the usefulness of a more complex model, it has to outperform the null model. Similarly, in order to justify a more complex metric, it has to perform better than \method{}.  

Like many non-trivial methods for caption evaluation,
\method{} does require an external resource for its definition.
However, rather than the resource being a trained parser, or a lexical database,
it is a set of pre-trained word embeddings.
Such pre-trained embeddings are publicly available for over 150 languages \cite{fasttext157lang}.
In general, this makes our proposed method more readily deployable than any other method requiring an external resource.

\section*{Acknowledgment}
We are grateful to Nvidia for providing Titan-Xp GPU, which was used for
our experiments. This work is supported by Australian Research Council, ARC DP150100294.

\end{document}